\documentclass{article}
\usepackage{arxiv}
\usepackage[utf8]{inputenc} 
\usepackage[T1]{fontenc}    
\usepackage{hyperref}       
\usepackage{url}            
\usepackage{booktabs}       
\usepackage{amsfonts}       
\usepackage{nicefrac}       
\usepackage{microtype}      
\usepackage{lipsum}		 
\usepackage[arabic,farsi,english]{babel}
\usepackage{graphicx}
\usepackage{amsmath}
\usepackage{subcaption}
\pdfoutput=1

\title{ACO-tagger: A Novel Method for Part-of-Speech Tagging using Ant Colony Optimization}
\date{March , 2023}
\author{ 
{Amirhossein Mohammadi} \\
	Department of Statistics Mathematics and Computer\\
	Allameh Tabataba'i University\\
	Tehran, Iran \\
	\texttt{\href{mailto:98.amirhosein@gmail.com}{98.amirhosein@gmail.com}} \\
	\And
 {Sara Hajiaghajani} \\
        Department of Statistics Mathematics and Computer\\
	Allameh Tabataba'i University\\
	Tehran, Iran \\
	\texttt{\href{mailto:sara.aghajani98@gmail.com}{sara.aghajani98@gmail.com}} \\
	\And
{Mohammad Bahrani} \\
        Department of Statistics Mathematics and Computer\\
	Allameh Tabataba'i University\\
	Tehran, Iran \\
	\texttt{\href{mailto:bahrani@atu.ac.ir}{bahrani@atu.ac.ir}} \\
}
\renewcommand{\shorttitle}{ACO-tagger: A Novel Method for Part-of-Speech Tagging using Ant Colony Optimization}

\begin{document}
\maketitle

\begin{abstract}
        Swarm Intelligence algorithms have gained significant attention in recent years as a means of solving complex and non-deterministic problems. These algorithms are inspired by the collective behavior of natural creatures, and they simulate this behavior to develop intelligent agents for computational tasks. One such algorithm is Ant Colony Optimization (ACO), which is inspired by the foraging behavior of ants and their pheromone laying mechanism. ACO is used for solving difficult problems that are discrete and combinatorial in nature. Part-of-Speech (POS) tagging is a fundamental task in natural language processing that aims to assign a part-of-speech role to each word in a sentence. This research paper proposed a high-performance POS-tagging method based on ACO called ACO-tagger. This method achieved a high accuracy rate of 96.867\%, outperforming several state-of-the-art methods. The proposed method is fast and efficient, making it a viable option for practical applications.
\end{abstract}


\section{Introduction}
In natural language processing, Part-of-Speech (POS) tagging is a task that involves identifying and labeling the parts of speech of words in a sentence, such as nouns, verbs, adjectives, adverbs, etc \cite{1}. This task is critical for disambiguating the meaning of words by understanding the structure and grammar of sentences. For example, the word "\FR{کاوش} " can refer to a name or an infinitive that means "exploration", and POS tagging can help differentiate which meaning is intended in a given context.
The purpose of POS tagging is to provide a standardized representation of a sentence, which can then be used as input for various NLP tasks, such as named entity recognition, syntactic parsing, and sentiment analysis \cite{2}. The goal of this task is to accurately assign a tag to each word in a sentence, which can be challenging due to the ambiguity and complexity of natural language \cite{3}.\\
POS tagging can also be used to extract more detailed information about a text, such as identifying named entities (e.g., people, places, and organizations) and detecting sentiment. It can also help improve the accuracy of language models and other NLP algorithms. Overall, the importance of POS tagging in NLP lies in its ability to provide a foundation for further language analysis and processing, enabling more sophisticated and accurate models to be built for a range of applications \cite{1}.\\
This article presented a new method for solving the POS-tagging task in NLP using the Ant Colony Optimization (ACO) algorithm, a novel approach that utilizes swarm intelligence and evolutionary computing algorithms. The ACO algorithm is employed as an optimizer to improve the tags on words, with the goal of finding the best tag sequence and providing the most optimal possible tag sequence as the final answer. The proposed method has shown promising results in improving the accuracy and efficiency of POS-tagging tasks. The research and experiments in this paper are based on the Persian language, using the Bijankhan linguistic corpus which is one of the most comprehensive linguistic corpora available in Persian. However, it should be noted that the algorithm is not dependent on the Persian language and can be adapted to work with other languages as well.\\
In order to assess the performance of ACO-tagger, we have compared it with the Viterbi algorithm. The Viterbi algorithm is a dynamic algorithm that uses Hidden Markov Models to solve problems, particularly in POS-tagging, by employing probabilistic models. This algorithm works by computing the probability of each possible sequence of tags and selecting the sequence with the highest probability as the output.
On the other hand, the ACO algorithm is a stochastic optimization method inspired by the behaviour of ants in nature that can solve non-linear problems. ACO works by simulating the behaviour of ants as they search for the shortest path between their nest and a food source. In the context of POS-tagging, the ACO algorithm is used to iteratively search for the best tag sequence for each word in a given sentence. By comparing the results of the ACO-tagger with those of the Viterbi algorithm, it is possible to determine whether the ACO approach is a viable alternative for POS-tagging tasks and to assess its strengths and limitations in comparison to other methods.

\section{Ant Colony Optimization}
Ant Colony Optimization (ACO) is a powerful metaheuristic method used in solving hard combinatorial optimization problems with discrete structures \cite{4}. ACO is inspired by the foraging behavior of ants, which involves a collective decision-making process. The algorithm was first presented by Dorigo et al. in 1999 and has since been widely used in various fields \cite{5}. ACO is known for its ability to find optimal solutions in complex problems by utilizing a stochastic process that involves simulating the foraging behavior of ants. In ACO, a colony of virtual ants is used to explore the search space, with each ant leaving pheromone trails that guide the search towards promising areas. Through this process, ACO is able to efficiently find near-optimal solutions to complex problems that are difficult to solve using traditional methods \cite{6}.\\
In nature, ants use a chemical substance called pheromone, specifically formic acid, to leave trails as they navigate. In the following, ants of the next generations can sense the amount of pheromone and use that information to follow the previously travelled paths. This process, known as marking, enables ants to achieve self-organization and find the optimal search path while foraging for food \cite{7}. By using this marking behavior, ants can explore a large area of their surroundings and converge toward the richest food sources. The algorithm uses pheromone trails to guide the search process and iteratively improves the solution by adjusting the pheromone levels on each candidate solution. The use of pheromone trails allows the ACO algorithm to efficiently explore the search space and converge to an optimal solution.\\
The algorithm operates by simulating the foraging behavior of ants, with each ant representing a potential solution to the optimization problem. In the beginning, each ant randomly navigates the search space, generating a path that represents its proposed solution. ACO incorporates pheromone trails into its search mechanism by assigning values to the edges of a graph to represent the strength of the pheromone trails. As ants follow the pheromone trails, the paths with stronger pheromone levels become more attractive and more likely to be selected by subsequent ants.\\
Similarly, the ACO algorithm assigns a probability to each edge based on the pheromone trail strength, and ants are more likely to choose edges with higher pheromone levels. The selected paths represent random variables, which are used to define a fitness function for each ant. The fitness function represents the objective function of the optimization problem, with a better fitness value indicating a shorter path and a better solution.\\
In the ACO algorithm, after the initial random paths are generated by the ants, the quality of the solutions provided by each ant is evaluated. Based on this evaluation, the amount of pheromone is updated for the paths taken by each ant. In the next iteration, the ants choose their paths randomly based on the available pheromone. The probability of choosing a particular path is calculated based on the amount of pheromone deposited on that path. The higher the amount of pheromone, the higher the probability of choosing that path. By iteratively updating the pheromone levels and re-generating the paths, the algorithm searches the solution space to find the optimal solution. This approach allows the algorithm to balance between exploration and exploitation of the solution space, which can result in finding better solutions. The amount of pheromone left by the ants on the path is updated according to the path that ant has passed. This means that the better solutions have a higher amount of pheromone, and as a result, in the next generations, more ants will choose the path with the higher amount of pheromone. This behavior leads to a positive feedback loop in which good solutions attract more ants, and as a result, the pheromone on that path increases even more \cite{8}.\\
In an experiment, Deneubourg was able to prove that ants will finally converge toward the shorter path. This experiment demonstrated that ants were capable of solving complex problems and finding the shortest path between their nest and a food source. The experiment also showed that ants are capable of self-organization and coordination, which is a fundamental principle of swarm intelligence and the basis for the development of the ACO algorithm \cite{9}. The effectiveness of this approach has been demonstrated in various applications, including job scheduling, network routing, and machine learning. In these applications, the ACO algorithm has been shown to be competitive with other state-of-the-art methods in terms of solution quality and computational efficiency.

\section{Previous related works}
Different methods have been used for tagging words, and the variety of methods is evident in this case.
The Brill Tagger method, a rule-based approach, was one of the first methods introduced for this work. In a study by Karbasian et al., the Brill Tagger method achieved a tagging accuracy of 94.27\% for the Persian language using the Bijankhan corpus \cite{11}. However, the challenge with rule-based approaches is that the rules for determining tags need to be defined and created, typically requiring an expert's assistance. As a result, different sets of rules may be developed based on the expert's perspective. \\
A different approach to tagging utilizes neural networks, specifically recurrent neural networks like LSTM. This method applies the effect of a word's category to the subsequent words in the recurrent layer, resulting in more accurate tagging. A study by Kochari and al. explored the Bijankhan corpus using LSTM, incorporating the previous word's tag (bigram), and achieved 88.94\% accuracy in correctly tagging words. In this method, by increasing the number of previous words, a higher percentage of accuracy can be achieved. As far as in the test with increasing words to 5 (5-gram), this algorithm has reached 95.6\% accuracy \cite{12}.\\
In a study on Bijankhan's corpus, Hosseini et al. utilized the artificial neural network (ANN) approach for POS tagging and achieved a high level of accuracy. Their research found that the ANN-based POS tagging system had an accuracy rate of 95.7\% \cite{13}.\\
One of the most famous and widely used methods used for POS tagging is HMM (Hidden Markov Model). This model chooses the most probable tag sequence as output by calculating the tag probability of each word. For this purpose, the words are considered as a chain of different states, and at each stage, the probability of each tag is calculated for the corresponding word. These calculations are based on the Markov chain method. Okhovvat et al. reported an accuracy of 98.1\% for the Hidden Markov Model in their study on the RCIS corpus \cite{14}. Azimizadeh et al. employed the HMM approach for POS tagging on the Bijan Khan corpus, which resulted in a 95.11\% accuracy rate in their research \cite{15}.\\
Besharati et al. carried out a research that integrated LSTM and HMM techniques to boost POS tagging. Their study accomplished remarkable outcomes, as they were able to attain a high level of accuracy of 97.29\% in tagging words \cite{16}.
In a different study, Bokaei et al. emphasized improving the Viterbi model in POS tagging. Through the introduction of a new method, they achieved an enhanced accuracy rate for the model, which demonstrates the potential for further development in this field of research \cite{17}.

\section{Methodology}
The Viterbi algorithm is a well-known approach for POS tagging, and it operates using the Markov model \cite{18}. Viterbi displays each word and sentence as a sequence of words and then calculates the score of each tag for each word according to the previous words and their scores by using emission and transition matrices. Finally, after calculating the scores for all tags, from the end of the sequence we choose the score that is maximized between the others for the last word and choose its tag then we continue backward until obtaining the first word’s tag. ACO-tagger is similar to Viterbi in that it views sentences as a sequence of words.\\
\begin{figure}
        \begin{center}
        \includegraphics[scale = 0.45, bb= 500 0 0 370]{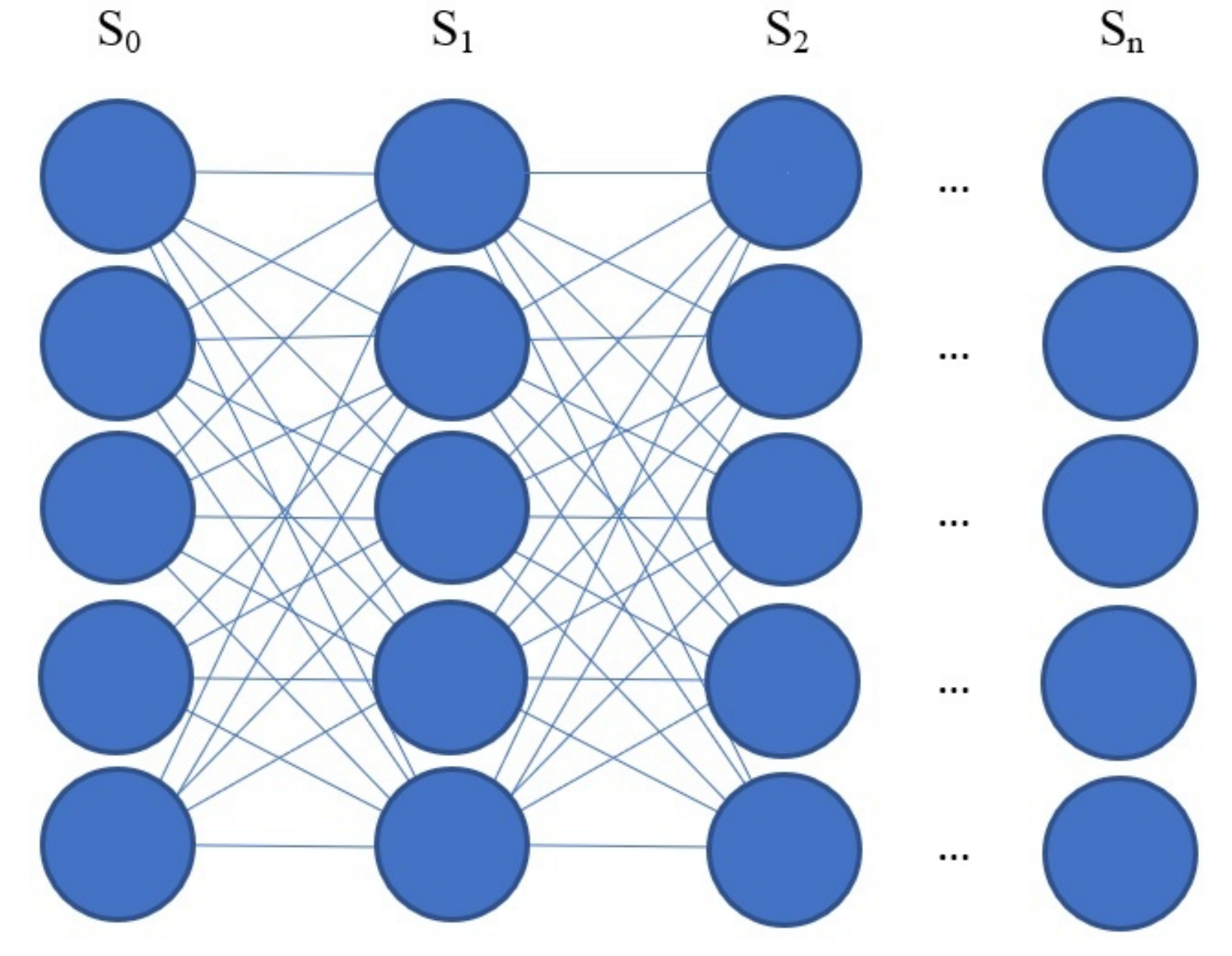}
        \end{center}
	\caption{The Trellis structure.}
	\label{fig:fig1}
\end{figure}
The trellis graph used in this method is composed of 'n' segments, each representing a word in the sentence, with nodes in each segment representing one of the possible tags for the corresponding word. In figure (1) the structure of the trellis and the connections between blocks and nodes has shown. This graph is formed by using emission and transition matrices. Distance between nodes is important in this graph so when we have a higher value for emission and transition matrices between two nodes, we should have a smaller distance between that two nodes in this graph.
\begin{itemize}
	\item Emission matrix: representation of the probability of assigning a specific POS tag to each word in the vocabulary.
	\item Transition matrix: representation of the probability of transitioning between POS tags in a sequence of words.
\end{itemize}
For determining the distance between two nodes, the following equation is used:
    \begin{equation}
        D_{i.j}=\left({emission}_{i_.j}\right)^{log{{(transition}_{i_.j}})}
    \end{equation}

As transition and emission values are in the type of probabilities, they will always be within the range of 0 to 1. The value of $\log(transition_{i,j})$ will be less than 0 due to this reason, and as the probability of $transition_{i,j}$ and $emission_{i,j}$ increases, the length of $D_{i,j}$ will become shorter.


When $emission_{i,j}$ is equal to zero, it indicates that the $tag_j$ is not associated with $word_i$. Similarly, if $transition_{i,j}$ is equal to zero , it indicates that $tag_j$ cannot appear after $tag_i$. To prevent assigning these tags to the words in the output response and potentially having impossible sequences of tags, their corresponding paths are assigned a cost of infinity. Hence, in equation (1) if ${emission}_{i_.j}=0$ or ${transition}_{i_.j}=0$ then the value of $D_{i_.j}$ is equal to $inf$. If the ant passes the path with the value of $inf$, the cost of the path traveled by that ant will always be equal to $inf$, and the answer produced by it will never be the final answer as an output.\\
After making the graph of a sentence, the next step is inserting ants into the graph. According to the ACO algorithm, ants randomly choose the ahead path based on a probability function. This probability is based on the pheromone on each branch. The following probability function is used to calculate the probability of choosing the path by the i-th ant:
\begin{equation}
    {P}_{i,d}\ =\ \frac{{c}_{{i},{d}}^{\alpha}\ .\ \eta_{{i},{d}}^{\beta}}{\sum_{{k = 1}}^{{B}}{({c}_{{i},{k}}^{\alpha}\ .\ \eta_{{i},{k}}^{\beta})}}     
\end{equation}
In this equation, the value $\eta$ is known as the heuristic value of the problem. A value is assigned to $\eta$ for each of the branches and paths. Also, $c_{i.j}$ expresses the amount of pheromone available in $branch_{i,j}$. At the beginning of the algorithm, the values of array $C$ are equal to zero because there is no pheromone on the branches of the graph. During the algorithm, depending on what the ants choose and where they move, these values are updated. The update takes place after all the ants of a generation have reached the last node, which is known as the food source. Changes in the pheromone of paths occur for two reasons:
\begin{enumerate}
    \item Reduction of available pheromone due to the evaporation over time
    \item The increase in pheromone due to the ant passing through it
\end{enumerate}
The amount of pheromone is updated at the end of each generation in the following way: 

\begin{equation}
c_{i,d}^{new}=\left(1-\rho\right)c_{i,d}\ + \sum_{j=1}^{M} \Delta c^{j}_{i,d}
\end{equation}
which $\rho$ is the evaporation rate of the pheromone.
At each stage, the ants of a generation choose the path and go through the nodes according to the amount of pheromone released by the previous generations. The best solution produced by each ant is equal to the shortest path traveled by them. Finally, after passing the path by all generations, the best solution produced is presented as the final solution.\\
In the following, we will discuss the algorithm in more detail by using an example. 
For example, the goal is to determine the tags for the words of the sentence "\FR{امروز هوا برفی است.}" where the table of emission, transition, and $\pi$ probabilities are shown in tables (1), (2), and (3) respectively.
\begin{table}
	\caption{Emission probabilities table}
	\centering
	\begin{tabular}{lccccc}
		\toprule

		&\FR{امروز}   & \FR{هوا} &\FR{برفی}  &\FR{است} &\FR{.} \\
            \cmidrule(r){2-6}
		N	&0.1	&1.0	&0.2	&0.0	&0.0     \\
		V	&0.0	&0.0	&0.0	&1.0	&0.0  \\
		ADJ	&0.0	&0.0	&0.8	&0.0	&0.0\\
            ADV	&0.9	&0.0	&0.0	&0.0	&0.0\\
            DELM	&0.0	&0.0	&0.0	&0.0	&1.0\\
		\bottomrule
	\end{tabular}
	\label{tab:table}
\end{table}
\begin{table}
	\caption{Transition probabilities table}
	\centering
	\begin{tabular}{lccccc}
		\toprule
		&N	&V	&ADJ	&ADV	&DELM \\
            \cmidrule(r){2-6}
		N	&0.6	&0.05	&0.2	&0.05	&0.2     \\
		V	&0.7	&0.1	&0.2	&0.0	&0.0  \\
		ADJ	&0.5	&0.0	&0.1	&0.15	&0.25\\
            ADV	&0.35	&0.05	&0.3	&0.1	&0.2\\
            DELM	&0.2	&0.7	&0.05	&0.05	&0.0\\
		\bottomrule
	\end{tabular}
	\label{tab:table}
\end{table}
\begin{table}
	\caption{${\pi}$ probabilities table}
	\centering
	\begin{tabular}{lccccc}
		\toprule
		&N	&V	&ADJ	&ADV	&DELM \\
            \cmidrule(r){2-6}
		Ø	&0.6	&0.01	&0.04	&0.3	&0.05     \\
		\bottomrule
	\end{tabular}
	\label{tab:table}
\end{table}

First, we should calculate the probabilities of tags for the word "\FR{امروز}" which comes as the first word in the sentence by using equation (1). The initial stage is illustrated in figure (2-a) by starting with $\emptyset$ state. The result of the calculations of this section is the cost of the initial edges. According to equation (1) and the explanations surrounding it, as well as the probabilities of table (1) and table (3), the length of the initial edges which are the edges between the initial state and the first word are calculated as follows.


\begin{figure}
\begin{subfigure}{.5\textwidth}
  \centering
  \includegraphics[width=.38\linewidth]{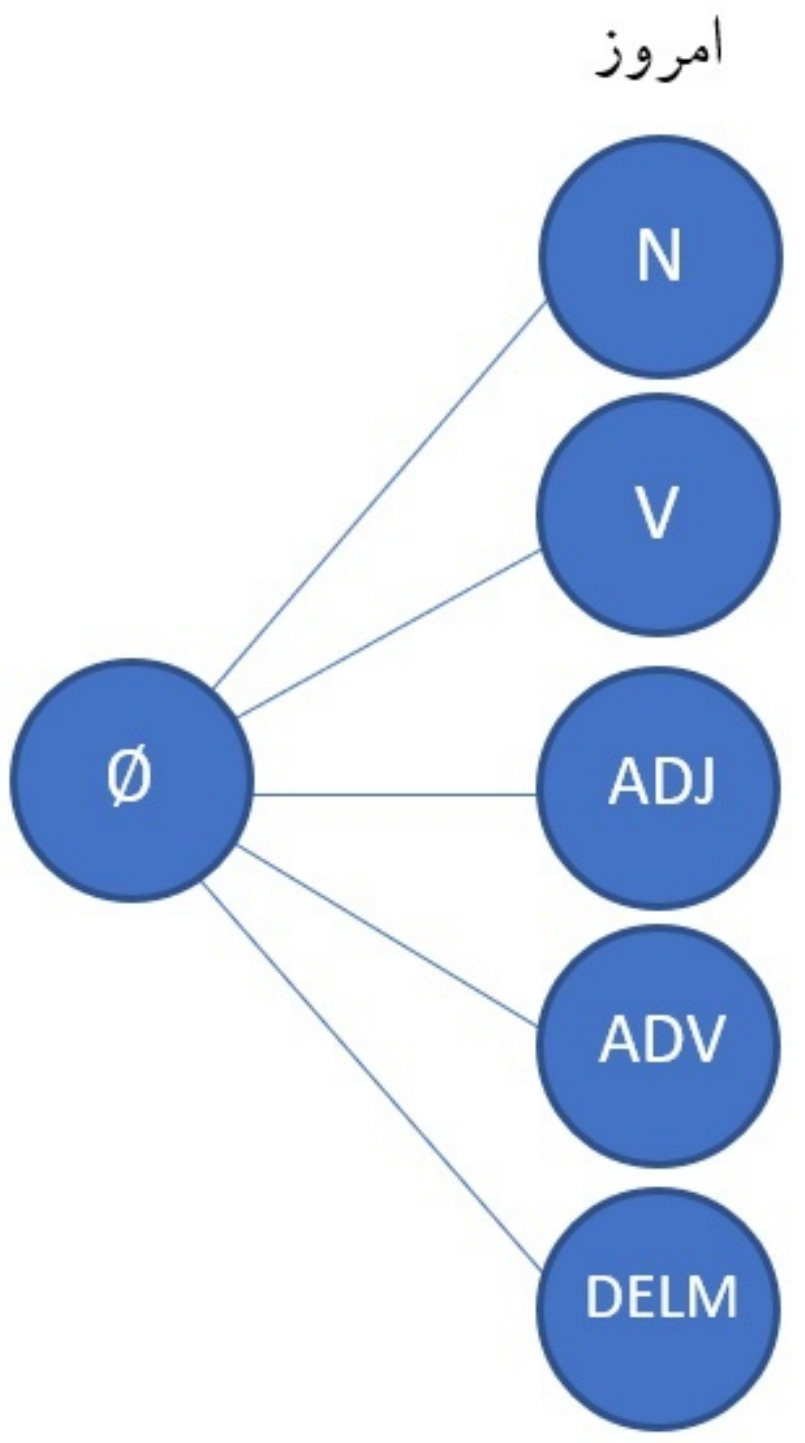}
  \caption{The initial state of choosing the path.}
  \label{fig:sfig1}
\end{subfigure}%
\begin{subfigure}{.5\textwidth}
  \centering
  \includegraphics[width=.4\linewidth]{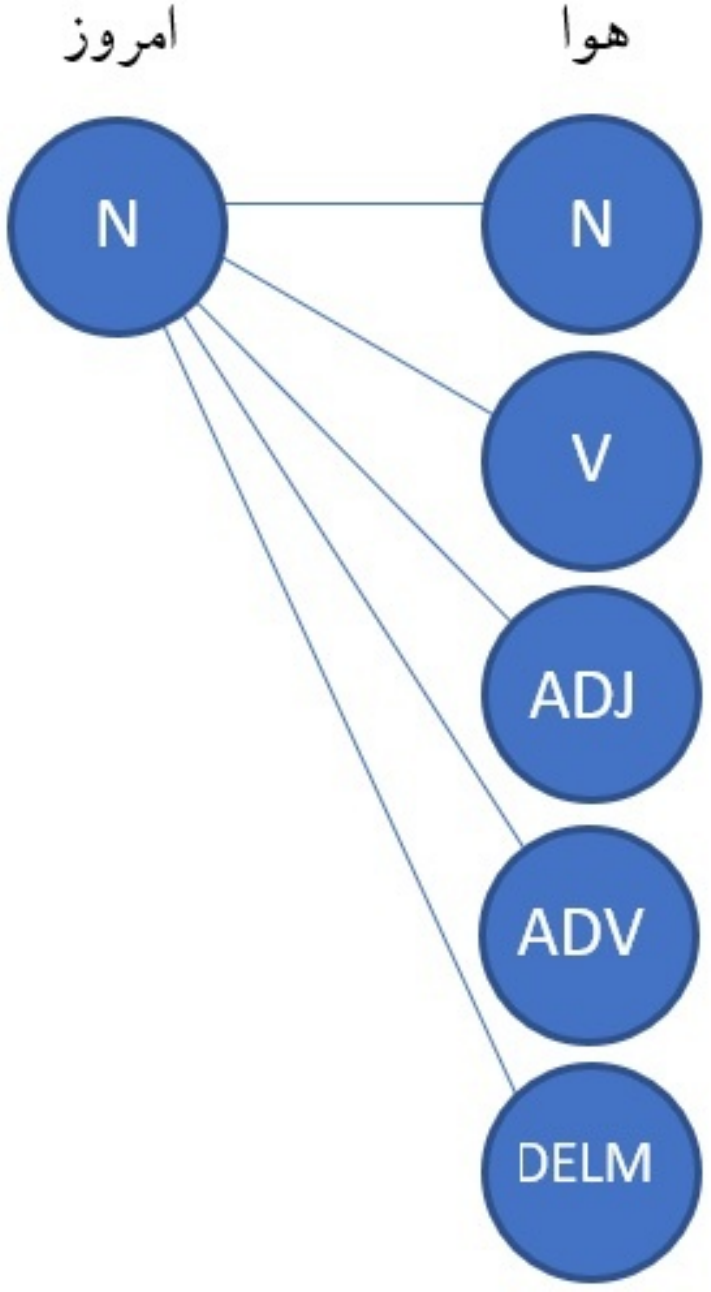}
  \caption{Paths between \FR{امروز} $\mid$ N  and the next state.}
  \label{fig:sfig2}
\end{subfigure}
\caption{Illustrations of the relation between statuses and conditions}
\label{fig:fig}
\end{figure}

$D(\emptyset, $\FR{امروز}$ \mid N) =  emission( $\FR{امروز}$ \mid N) ^ {log \pi (N)} = 1.667$ \\
$D(\emptyset, $\FR{امروز}$ \mid V) =  emission( $\FR{امروز}$ \mid N) ^ {log \pi (N)} = inf $ \\
$D(\emptyset, $\FR{امروز}$ \mid ADJ) =  emission( $\FR{امروز}$ \mid N) ^ {log \pi (N)} = inf$ \\
$D(\emptyset, $\FR{امروز}$ \mid ADV) =  emission( $\FR{امروز}$ \mid N) ^ {log \pi (N)} = 1.057$ \\
$D(\emptyset, $\FR{امروز}$ \mid DELM) =  emission( $\FR{امروز}$ \mid N) ^ {log \pi (N)} = inf$ \\

In the same way, the distance between other nodes is calculated, with the difference that in the following steps, instead of the $\pi$ value, the transition probabilities in the table (2) are used.Also, the connections between one of the tags in the first word and the second word's tags are shown in figure (2-b).

    
$D( $\FR{امروز}$ \mid N, $\FR{هوا}$ \mid N) = emission( $\FR{هوا}$ \mid N) ^ {log (transition (N|N))} = 1 $ \\
$D( $\FR{امروز}$ \mid N, $\FR{هوا}$ \mid V) = emission( $\FR{هوا}$ \mid V) ^ {log (transition (V|N))} = inf $ \\
$D( $\FR{امروز}$ \mid N, $\FR{هوا}$ \mid ADJ) = emission( $\FR{هوا}$ \mid ADJ) ^ {log (transition (ADJ|N))} = inf $ \\
$D( $\FR{امروز}$ \mid N, $\FR{هوا}$ \mid ADV) = emission( $\FR{هوا}$ \mid ADV) ^ {log (transition (ADV|N))} = inf $ \\
$D( $\FR{امروز}$ \mid N, $\FR{هوا}$ \mid DELM) = emission( $\FR{هوا}$ \mid DELM) ^ {log (transition (DELM|N))} = inf $ \\

To create attractiveness in choosing shorter paths compared to longer paths, in this method, we put the $\eta_{i.j}$ value of each path equal to $\frac{1}{D_{i.j}}$ . As a result of this work, the smaller the distance between two vertices, the larger the value of $\eta_{i.j}$ will be, so, the probability of choosing that path, which is calculated according to equation (2), will be higher. This work will make the length of the path effective in the possibility of choosing the path by each ant, and an ant will choose according to the length of the path and the amount of pheromone available on it when making a decision. \\
The length of the path traveled by each ant is stored in each traversal, and finally, the best answer provided (the shortest path traveled) by the members of each generation is compared with the best solution produced by the members of the previous generations, and the least expensive path traveled is saved as the optimal answer. The nodes traveled in this path are the most probable answers for POS tagging.\\
In the ACO algorithm, various stopping parameters can be set to complete the calculations. To use this algorithm for tagging words, the number of ants’ generations is considered as the stop parameter, and the algorithm will stop after all $N$ generations of ants passed through, and the best answer produced during these $N$ generations will be the output and shows the final tags. 

\section{Experimental Result}

The performance and efficiency of ACO algorithm's processing and computation depend directly on the parameters of the problem. Typically, this problem involves 6 primary variables:
\begin{itemize}
    \item \textbf{Generation:} This parameter in the ACO algorithm determines the number of ant generations. In this study, the number of generations is regarded as the stopping condition for the algorithm, which will halt after N generations. Increasing the number of generations can enhance the accuracy of the algorithm by enabling more ants to address the problem and emit pheromones. However, the trade-off is that the algorithm may run slower as the number of generations increases.

    \item \textbf{Ant Count:} The number of ants in each generation is controlled by this parameter. If the number of ants is too small, the algorithm may quickly converge to a suboptimal solution. Conversely, if the number of ants is too large, the algorithm may become slower.

    \item \textbf{$\alpha$:} the $\alpha$ parameter is a weight that balances the importance between the amount of pheromone left on the paths and the heuristic information, which provides a prior knowledge of the problem. A higher alpha value places more emphasis on the heuristic information, while a lower $\alpha$ value places more emphasis on the pheromone trails. In other words, $\alpha$ controls the degree to which the ants rely on the previous knowledge of the problem (heuristic information) or the information gained from their exploration (pheromone trail). The optimal value of $\alpha$ depends on the problem being solved and can be determined through experimentation. A higher $\alpha$ value is generally preferred for problems where prior knowledge plays an important role, while a lower $\alpha$ value may be better suited for problems where exploration is necessary.

    \item \textbf{$\beta$:} controls the weight given to the heuristic information about the problem being solved. This parameter determines the effect of the problem's domain knowledge on the movement of the ants. $\beta$ controls the relative influence of the pheromone trail and the heuristic information in the decision-making process of the ants. A higher value of $\beta$ gives more weight to the heuristic information, while a lower value of $\beta$ gives more weight to the pheromone trail. The optimal value of $\beta$ depends on the specific problem being solved and may require experimentation to find.

    \item \textbf{$\rho$:} the pheromone evaporation rate is controlled by this parameter. It determines how quickly the pheromone trail will evaporate over time. A high value of $\rho$ means that the pheromone evaporates quickly, while a low value of $\rho$ means that the pheromone trail will persist longer. By setting an appropriate value for $\rho$, the algorithm can balance the importance of the current and past solutions. If $\rho$ is set too high, the ants will tend to converge on a suboptimal solution too quickly, while setting $\rho$ too low may cause the ants to explore too much and take a longer time to find a good solution. Therefore, the value of $\rho$ must be chosen carefully based on the characteristics of the problem being solved.

    \item \textbf{Pheromone Quantity:} The pheromone quantity parameter in ACO algorithm has a dual effect on the algorithm. First, it helps ants to find potential solutions by creating a stronger and longer-lasting pheromone trail, but a very high quantity can trap the algorithm in local optima. Second, a high pheromone quantity can cause fast convergence of the algorithm, leading to suboptimal solutions. To prevent this, it is essential to balance the pheromone quantity with other parameters, including the exploration-exploitation trade-off and the pheromone evaporation rate.

    The experimental results indicate that if the pheromone does not evaporate, it can adversely affect the performance of the algorithm for POS-tagging. Furthermore, the number of ants and generations should be fine-tuned to achieve both acceptable accuracy and desirable execution time. Table (4) presents the optimal combination of these parameters obtained through numerous experiments. To determine this combination, all feasible parameter combinations were tested within the designated intervals.
\end{itemize}

\begin{table}
	\caption{Input parameters}
	\centering
	\begin{tabular}{cccccc}
		\toprule
		Generation   &Ants    &$\alpha$   &$\beta$   &$\rho$  &quantity \\
            \midrule
		3 &20 &0.9    &0.9    &0.95   &10     \\
		\bottomrule
	\end{tabular}
	\label{tab:table}
\end{table}

\section{Conclusion}
The experiments for this method utilized the Bijankhan corpus, which is a vast collection of text in the Persian language consisting of 2.6 million individual words categorized into 32 different tags, covering 4,300 diverse topics \cite{19}. During the experiments, the training data for the method was composed of 80\% of the sentences in the corpus, which were used to create probability matrices. The remaining 20\% of the data was used as the test data.\\
As a means of comparing the ACO-tagger to the Viterbi algorithm in a more fair and accurate manner, both algorithms have been tested concurrently. Both algorithms were used to tag each sentence in the test data, and the final accuracy of both algorithms was calculated on the same dataset. The results of the comparison showed that the ACO-tagger achieved an accuracy of 96.867\%, which is higher than the Viterbi algorithm's accuracy of 96.361\%. Additionally, the ACO-tagger demonstrated superior performance in tagging lengthy sentences compared to Viterbi, which accuracy decreases as the sentence length increases.\\
The findings from the conducted tests suggest that the ACO-tagger is a promising method for sequence labeling tasks such as POS-tagging. Compared to previous methods, ACO-tagger exhibited superior performance in terms of accuracy, especially for lengthy sentences. This indicates that the algorithm is robust and effective in solving complex optimization problems. Moreover, ACO-tagger's ability to generalize across different datasets makes it a versatile and adaptable tool for various applications. Its potential is not limited to language processing but can also be extended to other domains, such as speech recognition, image segmentation, and bioinformatics. Overall, the results indicate that ACO-tagger is a powerful and promising method for optimizing sequence labeling tasks, and it opens up new possibilities for future research in this field.\\
In addition to ACO, other swarm intelligence algorithms can also be applied for optimization problems in various fields. Among these algorithms, the Firefly Algorithm (FA) \cite{20} and Artificial Bee Colony Algorithm (ABC) \cite{21} are promising approaches. The FA algorithm is inspired by the flashing patterns of fireflies and has shown success in solving optimization problems such as feature selection, image segmentation, and clustering. Similarly, the ABC algorithm is based on the foraging behavior of honey bees and has demonstrated strong performance in optimizing various functions, including feature selection, data classification, and image processing. Overall, swarm intelligence algorithms such as ACO, FA, and ABC present powerful and versatile tools for solving optimization problems in diverse applications.

\bibliographystyle{unsrt}

\end{document}